\documentclass[11pt]{article}
\usepackage[
  a4paper,
  textwidth=16cm,
  textheight=24cm,
  centering
]{geometry}

\usepackage{graphicx}
\usepackage{amsmath}
\usepackage{amssymb}
\usepackage{amsthm}
\usepackage{multicol}
\usepackage{multirow}
\usepackage{subcaption}
\usepackage{url}
\usepackage[mathscr]{euscript}
\usepackage{array}
\usepackage{float}
\usepackage{xspace}
\usepackage{comment}
\usepackage{tabularx}
\usepackage{todonotes}
\usepackage{bm}
\usepackage{enumitem}
\usepackage{hyperref}
\usepackage{tikz}
\usepackage{thmtools}
\usepackage{mathrsfs}
\usepackage{thm-restate}
\usepackage{stmaryrd,nicefrac,mathtools}

\usepackage{amsthm,amsmath,amssymb,amsfonts,multicol}
\usepackage{enumitem,caption}

\usepackage{graphicx} 
\usepackage{tikz}
\usepackage{xcolor}
\usepackage{amsmath}
\usepackage{amssymb}
\usepackage{amsthm}
\usepackage{multicol}
\usepackage{multirow}
\usepackage{caption}
\usepackage{stmaryrd}
\usepackage{array}
\usepackage[mathscr]{euscript}
\usepackage{xspace}
\usepackage{float}
\usepackage{mathtools}
\usepackage{enumitem}
\usepackage{hyperref}
\usepackage{booktabs}
\usepackage{colortbl}
\usepackage{tcolorbox}

\newcommand{\zee}{\mathsf{Z}\,}

\newcommand{\condition}{{\sf Cond}}

\newcommand{\response}{{\sf resp}}

\newcommand{\last}{{\sf LAST \xspace}}
\newcommand{\scope}{{\sf scope \xspace}}

\newcommand{\stopCondition}{{\sf stopC}}

\newcommand{\fret}[5]{\textcolor{red}{#1} \textcolor{yellow}{#2} \textcolor{green}{#3} shall \textcolor{cyan}{#4} \textcolor{violet}{#5} }

\usetikzlibrary{decorations.pathmorphing}

\usetikzlibrary{positioning}
\usetikzlibrary {shapes.misc}
\usetikzlibrary {decorations.pathreplacing, calligraphy}
\usepackage{standalone} 
\usepackage{tikzscale} 

\definecolor{steelblue}{rgb}{0.27, 0.51, 0.71}

\newcommand{\fretish}{\textsc{Fretish}\xspace}
\newcommand{\Fret}{\textsc{Fret}\xspace}
\newcommand{\FV}{\textit{Fretish-to-MTL}\textsubscript{FV}\xspace}
\newcommand{\NASA}{\textit{Fretish-to-MTL}\textsubscript{FRET}\xspace}

\definecolor{rowgray}{gray}{0.93}
\definecolor{headerblue}{RGB}{220,230,242}

\newcommand{\U}{\mathbin{\mathsf{U}}}
\newcommand{\Since}{\mathbin{\mathsf{S}}}
\newcommand{\R}{\mathbin{\mathsf{R}}}
\newcommand{\X}{\mathsf{X}\,}
\newcommand{\F}{\mathsf{F}\,}
\newcommand{\On}{\mathsf{O}\,}
\newcommand{\G}{\mathsf{G}\,}
\newcommand{\Y}{\mathsf{Y}\,}
\newcommand{\Z}{\mathsf{Z}\,}

\newcommand{\Hist}{\mathsf{H}\,}

\newcommand{\Ub}[1]{\mathbin{\mathsf{U}_{[#1]}}\,}
\newcommand{\Fb}[1]{\mathsf{F}_{[#1]}\,}
\newcommand{\Gb}[1]{\mathsf{G}_{[#1]}\,}

\newcommand{\Us}{\mathbin{\mathsf{U}_{\phi_s}}}

\newcommand{\exitS}{\sf exit_s}

\newcommand{\sameInf}{\textit{= Fut. Inf.}\xspace}
\newcommand{\Only}{\textsc{Only}\xspace}
\newcommand{\NotOnly}{\textsc{Reg}\xspace}

\newcommand{\exitWithin}[1]{\phi_s \Ub{0,#1} \neg \phi_s}

\usepackage[table]{xcolor}

\definecolor{rowgray}{gray}{0.92}
\definecolor{headerblue}{RGB}{220,230,240}
\definecolor{blockgray}{gray}{0.85}

\usepackage{authblk}

\begin{document}

\title{Coherency through formalisations of Structured Natural Language, A case study on \fretish.}

\author[1,2]{Joost J. Joosten}
\author[3]{Marina López Chamosa}
\author[1,3]{Sofía Santiago Fernández}

\affil[1]{\small Universitat de Barcelona, Barcelona, Spain}
\affil[2]{\small Centre de Recerca Matemàtica, Barcelona, Spain}
\affil[3]{\small Formal Vindications S.L., Barcelona, Spain}

\date{}

\maketitle

\begin{abstract}
Formalisation is the process of writing system requirements in a formal language. These requirements mostly originate in Natural Language. In the field of Formal Methods, formalisation is often identified as one of the most delicate and complicated steps in the verification process. Not seldomly, formalisation tools and environments choose various levels of requirement descriptions: Natural Language, Technical Language, Diagram Representations and Formal Language, to mention a few. In the literature, there are various maxims and principles of good practice to guide the process of requirement formalisation. In this paper we propose a new guideline: Coherency through Formalisations. The guideline states that the different levels of formalisation mentioned above should roughly follow the same logical structure. The principle seems particularly relevant in the setting where LLMs are prompted to perform reasoning tasks that can be checked by formal tools using Structured Natural Language to act as an intermediate layer bridging both paradigms. In the light of coherency, we analyze NASA’s Formal Requirement Elicitation Tool FRET and propose an alternative automated translation of the Controlled Natural Language \fretish to the formal language of MTL. We compare our translation to the original translation and prove equivalence using model checking. Some statistics are performed which seem to favor the new translation. As expected, the translation process yielded interesting reflections and revealed inconsistencies which we present and discuss.
\end{abstract}

\section{Introduction}

The formalisation of system requirements is widely seen as an important bottleneck in the field of formal methods \cite{Rozier:2016:SpecificationBottleneck}. The machine can only process rigid formal syntax that is difficult to parse and understand for the human agent, while natural language is often not precise enough (open texture \cite{GuittonEtAl:2025:OpenTextureInLaw}), prone to ambiguity and in need of contextual information or common knowledge that may vary over time, location, legislation, etc. 

This tension places the quest for suitable languages at a central place within formal methods and there is a long tradition, both in industry and academia, of languages and tools to bridge the gap between natural languages and machine languages, especially in the realm of temporal requirements that shall be the focus of this paper \cite{mueckstein:1985:CNL, GiannakopoulouEtAl:2020:FormalRE, DwyerEtAl1999SPS, MenghiEtAll:2019:SpecificationPattersRoboticMissions, HuttnerMerigoux:2022:CatalaFuture,  FifarekEtAl:2017:Spear}.

However, unambiguous, yet understandable meaning is not the only objective of good requirement assertion languages. For example, the tension between expressiveness on the one hand and computational complexity on the other, is another important guideline in the quest for optimal languages for requirement formulation \cite{mullerJoosten:2023:modelcheckingPreprint, BogliEtAl:2025:NaturalFormalisations, BrunelloEtAl:2019:SynthesisOfLTLformulas}. There is an extensive corpus of literature that reflects on the nature of good  formalisations taking all sorts of qualities into account \cite{BogliEtAl:2025:NaturalFormalisations, Buzhinsky:2019:formalization, mullerJoosten:2023:modelcheckingPreprint, PeregrinEtAl:2013:Criteria} and this paper contributes to this body of work as follows. 

In Section \ref{section:principlesOfGoodFormalization} we propose and discuss the new quality of \textit{coherency} of formalisations which is roughly that formalisations come with different layers of abstraction and that all these layers follow the same approximate logical structure. 
To showcase the new principle, we first revisit various fragments of the temporal logic MTL in Section \ref{section:TemporalLogics}, describe NASA's requirement language \fretish in Section 
\ref{section:FRET} to present a new translation of \fretish into MTL in Section \ref{section:Translations}. Then, in Section \ref{section:Comparison} we compare the new translation to NASA's translation in the light of Coherency and also compare along other metrics like length and modal nesting. The final section contains some concluding reflections and ideas for future research.

\section{Principles of Good Formalisation}\label{section:principlesOfGoodFormalization}

Building on \cite{BogliEtAl:2025:NaturalFormalisations}, we adopt and restate three criteria (numbering is ours) for what the authors call a \emph{natural} formalisation:

\begin{itemize}
\item[(N1)]\label{naturallnessCondition1}
The formalisation should be driven by the temporal structure of the requirement itself, rather than by a priori commitment to a specific logic.

\item[(N2)]\label{naturallnessCondition2}
The chosen logic should be minimal, providing just enough expressive power to capture the requirement, thus prioritizing adequacy over expressiveness.

\item[(N3)]\label{naturallnessCondition3}
Formalisations should be as compact as possible, provided that compactness does not compromise readability and intuitive understanding.
\end{itemize}

These criteria focus on the quality of individual formalisations. However, they do not explicitly address how different representations of the same requirement should relate to one another. In practice, requirements are rarely expressed in a single form: they typically evolve across multiple layers, from informal descriptions to structured specifications and formal models.

To address this gap, we introduce a complementary principle, which we call \emph{coherency}.

\medskip

\noindent\textbf{Logical structure.}
We use the term \emph{logical structure} to denote the abstract compositional pattern underlying a requirement, independently of the specific language in which it is expressed. In our setting, this typically corresponds to a decomposition into components such as \emph{scope}, \emph{condition}, and \emph{timed response}, combined through a fixed pattern (e.g., trigger $\rightarrow$ obligation under a temporal context).

This notion is closely related to the idea of \emph{structure-preserving translations} and \emph{refinement} in formal methods, where mappings between representations aim to preserve semantic relationships and compositional organization (see, e.g., \cite{mavridou2023formal} and classical notions of refinement in system design).

\medskip

\noindent\textbf{Coherency.}
We define coherency through two principles:

\begin{itemize}
\item[(C1)] It is good practice to structure system requirements across multiple layers of formalisation, such as natural language, structured or technical language, diagrammatic representations, and formal specifications (possibly enriched with macros).

\item[(C2)] It is good practice for these layers to share a common logical structure, so that each layer can be understood as a refinement of the previous one, rather than an independent reformulation.
\end{itemize}

The notion of \emph{refinement} plays a central role in this context: each layer should preserve the intent of the previous one while making certain aspects more precise or more amenable to formal analysis. Coherency thus ensures that moving between layers does not introduce unintended discrepancies, but instead supports a systematic and traceable development of requirements.

Importantly, coherency complements (N1)--(N3): while naturalness emphasizes adequacy, minimality, and readability within a given formalisation, coherency emphasizes alignment and traceability \emph{across} formalisation layers.

\paragraph{Discussion and potential limitations.}
While coherency provides a useful guiding principle, both (C1) and (C2) raise potential concerns.

Regarding (C1), introducing multiple layers of formalisation may increase the overall complexity of the development process. Maintaining consistency across natural language, structured representations, and formal specifications can be costly and error-prone. In practice, this may lead to redundancy, where similar information is expressed in multiple forms without clear added value. Moreover, not all applications require such a rich stratification: for simpler systems, a single formal layer may suffice, making additional layers unnecessarily burdensome.

Regarding (C2), enforcing a shared logical structure across all layers may restrict expressiveness. Natural language and diagrammatic representations often rely on abstractions or implicit context that do not map cleanly to formal logic. Imposing a uniform structure may therefore lead to oversimplifications or force unnatural encodings. Furthermore, different stakeholders (e.g., engineers, domain experts, verification specialists) may prefer different representations optimized for their specific tasks, rather than a single coherent structure.

The notion of refinement also introduces challenges. While refinement ideally preserves intent across layers, ensuring semantic equivalence is non-trivial. Subtle mismatches may arise, especially when moving from informal to formal representations, potentially undermining the very coherency the approach aims to guarantee. In addition, refinement relations may not always be strictly hierarchical; in some cases, different layers capture complementary rather than strictly refined views.

Despite these limitations, we argue that coherency remains a valuable design principle. The goal is not to enforce strict uniformity but to promote traceability and alignment across representations, enabling more systematic reasoning about requirements and their formalisations. We will showcase how the principle of Coherency can be respected for a translation of NASA's formal requirement language \fretish into various fragments of the temporal logic MTL which we will introduce in the next section.

\section{Preliminaries on Metric Temporal Logic}\label{section:TemporalLogics}
We shall be working with \textit{Linear Temporal Logic} (LTL, \cite{Pnueli:1977:LTL,Prior:1957:TimeAndModality}) and its variant Metric Temporal Logic (MTL, \cite{Koymans:1990:MTL, AlurHenzinger:1991:RealTimeLogics}) where temporal operators are restricted to intervals. We define the set $\mathbb I$ of intervals and their members as follows:
\begin{equation*}
\begin{split}
& \mathbb I : = \ \{ [a,b] \mid  a ,b \in \omega \mbox{ with } a\leq b \} \cup \{ [0,\infty) \} \ \ \mbox{ and } \Big(c\in [a,b] \Leftrightarrow a\leq c \leq b \Big) \\ & \mbox{ and } \Big(\forall\, n{\in} \omega \ n \in [0,\infty) \Big).
\end{split}
\end{equation*}

Throughout the paper, $\sf Prop$ denotes a countable set of propositional variables. Formulas shall be evaluated on so-called \textit{traces}. Traces can be finite or infinite and both are denoted mostly by $\rho$. An infinite trace is a function from the natural numbers to sets of propositional variables, that is, $\rho : \omega \to \mathcal P ({\sf Prop})$. Thus, $\rho(i)$ denotes a set of propositional variables for each $i \in \omega$: the propositional variables true at time $i$. For infinite traces $\rho$ we define the length $|\rho|$ to be $\omega$ with the convention that $i<\omega$ for any $i\in \omega$.

Finite traces $\rho$ have finite positive length $|\rho|$ and are maps from initial segments of the natural numbers to sets of propositional variables. That is, $\rho : \{0, \ldots, |\rho|-1\} \to \mathcal P ({\sf Prop})$. Our definitions below apply to traces of both finite and infinite length. In applications, the context should make clear if traces are assumed to be of finite or of infinite length.

We follow tradition and first present past and future time fragments of temporal logics separately. 
\subsection{Past time metric temporal logic}

We define the logic pMTL using the following syntax where $I$ is an interval in $\mathbb I$:
\[
\varphi := {\sf Prop}  \mid \bot  \mid (\varphi \to \varphi)    \mid \Y \varphi  \mid  \On_I \varphi \mid \Hist_I \varphi \mid (\varphi \Since_I \varphi)  \mid \Z \varphi.
\]
Since we work over classical logic, we define the other Boolean connectives as usual $\neg \varphi := (\varphi \to \bot)$, $\varphi \wedge \psi := \neg (\varphi \to \psi)$ and $\varphi \vee \psi:= \big( (\neg \varphi) \to \psi \big)$. For $J=[0,\infty)$ we will simply write $\On$, $\Hist$ and $\Since$ for  $\On_J$, $\Hist_J$ and $\Since_J$ respectively.
The modal operators $\Y$, $\On$, $\Hist$, and $\Since$ represent the temporal operators \textit{Previous} (\textit{Yesterday}), \textit{Once}, \textit{Historically}, and \textit{Since} , respectively. In addition, we consider the temporal operator $\zee$ which is like a relaxed version of $\Y$. The precise behavior of the operators is captured by the forcing relation $\Vdash$ where $\nVdash$ denotes the negation of $\Vdash$ and where quantification ranges over natural numbers: 
\[
\begin{array}{lll}  
\rho, t \nVdash \bot  & \mbox{ for all $t\in \omega$;}\\     

\rho, t \Vdash P  & \mbox{ if and only if } & P\in \rho (t) \ \ \mbox{for $P\in {\sf Prop}$};\\     

\rho, t \Vdash \varphi \to \psi  & \mbox{ if and only if } & \rho, t \nVdash \varphi \ \ \mbox{ or } \rho, t \Vdash \psi \\
\rho, t \Vdash \Y \varphi  & \mbox{ if and only if } & t>0 \ \wedge \ \rho, (t-1) \Vdash  \varphi;\\

\rho, t \Vdash \Z \varphi  & \mbox{ if and only if } & t=0 \vee (t>0 \ \wedge \ \rho, (t-1) \Vdash  \varphi);\\

\rho, t \Vdash \On_I \varphi  & \mbox{ if and only if } & \exists \, t_0 {\leq} t \Big( (t-t_0) \in I \ \wedge \ \rho,{t_0} \Vdash \varphi \Big);\\

\rho, t \Vdash \Hist_I \varphi  & \mbox{ if and only if } & \forall \, t_0 {\leq} t \Big( (t-t_0) \in I \ \to \ \rho,{t_0} \Vdash \varphi\Big);\\

\rho, t \Vdash \varphi \Since_I \psi & \mbox{ if and only if } & \exists \, t_0 {\leq} t \ \Big((t-t_0) \in I \ \wedge \ \rho,{t_0} \Vdash \psi \wedge \\ & & \forall t_1 \ (t_0{<}t_1{\leq} t \to \rho , {t_1} \Vdash \varphi) \Big) .\\
\end{array}
\]

We observe that the $\Y$ operator is intuitive except at the starting point. For example, for $t>0$ we have $\rho, t \Vdash \neg \Y \varphi \ \leftrightarrow \Y \neg \varphi$ and also $\rho, t \Vdash \Y (\varphi\to \psi) \to (\Y \varphi \to \Y \psi)$. However, for $t=0$ we have $\rho, 0 \Vdash \neg \Y \varphi$ for any formula $\varphi$ and, in particular $\rho, 0 \Vdash \neg \Y \top$ where $\top$ is short for $(\bot \to \bot)$. Consequently, we can define the starting point of any trace $\rho$ via $\rho, t \Vdash \neg \Y \top \ \ \Longleftrightarrow \ \ t=0$.

We will write $\rho \models \varphi$ to denote $\forall t\ \rho, t \Vdash \varphi$. Thus, the difference between $\zee$ and $\Y$ is that for all $\rho$,  $\rho \models \Z \top$ while $\rho \not \models \Y \top$, whereas  for all $t>0$, indeed $\rho, t \Vdash (\Z \top \wedge \Y \top)$.

The logic pLTL is the fragment of pMTL where no intervals are mentioned. By our reading convention, we can equivalently state that pLTL is the fragment of pMTL where every interval mentioned is equal to $[0,\infty)$. Although in finite and infinite words in real-time full MTL is undecidable \cite{OuaknineAndWorrell:2006:MTLUndecidable}, in discrete time, pMTL is just a succinct representation of pLTL.

The semantic definition of $\Since_I$ has a quantifier alternation: first an existential quantifier, followed by a bounded universal one. 
We observe that only the existential quantifier is restricted to the interval $I$. Notwithstanding, we still have certain interdefinability like $\models \On_I \varphi \ \leftrightarrow \ (\top \Since_I \varphi)$.

\subsection{Future time metric temporal logic}

For the logic fMTL we use the following syntax:
\[
\varphi := {\sf Prop}  \mid \bot   \mid (\varphi \to \varphi)   \mid \X \varphi  \mid  \F_I \varphi \mid \mathcal \G_I \varphi \mid (\varphi \U_I \varphi) \mid (\varphi \R_I \varphi)
\]
We will use the same writing conventions from pMTL also for fMTL. The modal operators $\X$, $\F$, $\G$, $\U$ and $\R$  stand for \textit{Next}, \textit{Future}, \textit{Globally}, \textit{Until} and \textit{Release} respectively and their semantic behavior is defined as
\[
\begin{array}{lll}
\rho, t \Vdash \X \varphi  & \mbox{ if and only if } & (t +1) < |\rho| \ \wedge \ \rho, (t+1) \Vdash  \varphi;\\

\rho, t \Vdash \F_I \varphi  & \mbox{ if and only if } & \exists \, t_0 {\geq} t \Big( (t_0-t) \in I \ \wedge \ \rho,{t_0} \Vdash \varphi \Big);\\

\rho, t \Vdash \G_I \varphi  & \mbox{ if and only if } & \forall \, t_0 {\geq} t \Big( (t_0-t) \in I \ \to \ \rho,{t_0} \Vdash \varphi\Big);\\

\rho, t \Vdash \varphi \U_I \psi & \mbox{ if and only if } & \exists \, t_0 {\geq} t \ \Big((t_0-t) \in I \ \wedge \ \rho,{t_0} \Vdash \psi \wedge \\ & & \forall t_1 \ (t{<}t_1{\leq} t_0 \to \rho , {t_1} \Vdash \varphi) \Big);\\

\rho, t \Vdash \varphi \R_I \psi & \mbox{ if and only if } & \forall\, t_0{\geq} t \ \Big( (t_0 -t) \in I \to \rho, t_0 \Vdash \psi \Big)\ \bigvee \ \\
 & & \exists t_0 \geq t \Big( (t_0 - t) \in I \wedge \rho, t_0 \Vdash \varphi \wedge \\ & & \forall t_1 \ \big( (t\leq t_1 \leq t_0 \wedge (t_1 - t) \in I) \to \rho, t_1 \Vdash \psi \big)\Big) .\\
\end{array}
\]
With these definitions the usual dualities like $A \ \R_I \  B \ \ :\Longleftrightarrow \ \ \neg (\neg A\ \U_I \ \neg B)$ remain valid as usual.
The logic fLTL is the fragment of fMTL where no intervals are mentioned. Again, by our reading convention we can equivalently state that fLTL is the fragment of fMTL where every interval mentioned is equal to $[0,\infty)$.

\subsection{Mixed Time Temporal Logic}

One can also consider temporal logics with both future and past time and this is the default choice that we make in this paper. The resulting syntax is
\[
\varphi := {\sf Prop}  \mid \bot  \mid (\varphi \to \varphi)    \mid \Y \varphi  \mid  \On_I \varphi \mid \Hist_I \varphi \mid (\varphi \Since_I \varphi)  \mid \Z \varphi \mid \X \varphi \mid (\varphi \U_I \varphi) \mid (\varphi \R_I \varphi).
\]
The semantic definitions of the operators is as above.
It is known that the expressibility of the resulting logic MTL does not change with respect to its past or future time restrictions (\cite{GabbayEtAl:1980:TemporalAnalysisFairness}), however, certain formulas are known to become exponentially more succinct \cite{Markey:2003:TemporalWithPastMoreSuccinct}.

\section{NASA's Formal Requirement Elicitation Tool (FRET)}\label{section:FRET}

NASA has developed an open source tool FRET: Formal Requirement Elicitation Tool. An important functionality is to translate a specifically designed structured natural language, called \fretish, to various fragments of the formal language of MTL. In this section we describe the features of \fretish and mention some other functionalities of FRET.

\subsection{The \fretish language}

\fretish shares features of both natural language and formal language. As such, it has a rather limited grammatical structure and limited vocabulary. All sentences in \fretish have the following form
FRETISH requirements follow the format:

    \begin{figure}[!h]
    \centering
    \includegraphics[width=\linewidth]{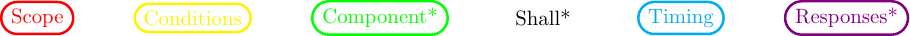}
    \end{figure}
The three fields marked with a star ($\star$) are mandatory while the other three fields are optional. A running example from \cite{ConradEtAl:2022:CompositionalProofFrameworkFretish} is 
\begin{quote}
    \fret{In $\sf flight mode$}{when $\sf horizontal\_distance$ $\leq$ 250 \& $\sf vertical\_distance$ $\leq$ 50}{the aircraft}{within 3 seconds}{satisfy $\sf warning\_alert$}
\end{quote}

In this paper and in other treatments \cite{GiannakopoulouEtAl:2020:FormalRE, Giannakopoulou:2021:AutomatedFormalizationOfSNLReqs, ConradEtAl:2022:CompositionalProofFrameworkFretish}, conditions like ``$\sf horizontal\_distance$ $\leq$ 250" are treated as boolean conditions and the semantics cannot reason about the $\leq$ order. In particular, statements like ``$\sf horizontal\_distance$ $\leq$ 250" and ``$\sf horizontal\_distance$ $>$ 300" will be treated as two independent propositional variables. When \fretish is embedded in a richer context like Requirement-Based Testing \cite{PecheurEtAl:2009:AnalysisReqBasedTesting, KatisEtAl:2025:StreamlinedReqBasedTesting} or Realizability Checking \cite{KooiMavridou:2019:RealizabilityCheckingInFRET, KatisEtAl:2022:RealizabilityCheckingInFRET, GiannakopoulouEtAl:2021:RealizabilityWithinFRET} both layers of semantics do interact.

In the example above, also $\sf flightmode$ and $\sf warning\_alert$ are boolean variables that can be true or false in various discrete moments of time. Thus, requirements are evaluated over discrete traces that determine the relevant boolean variables over time. 
The language \fretish evolves over time. We have based our current paper on the set of attributes studied in \cite{ConradEtAl:2022:CompositionalProofFrameworkFretish}. Thus, we consider 
\begin{itemize}
    \item 
    Eight scopes: In, Not In, Only In, Before, Only Before, After, Only After and, the Global scope.
    \item 
    Ten timings: Immediately, Eventually, Next, Always, Never, Within$[0,k]$, For$[0,k]$, After$[0,k]$, Until $\sf Stop$, Before $\sf Stop$ 
\end{itemize}  
Here, $\sf Stop$ is a reserved propositional variable to indicate a stop condition and $k$ is a natural number.
In \cite{ConradEtAl:2022:CompositionalProofFrameworkFretish} they considered two condition options: either a condition or no condition at all. This gave rise to $8\times 2 \times 10 = 160$ different templates (disregarding the parameter $k$). In a later version of \fretish there is a distinction between so-called \textit{continual} and \textit{trigger} conditions. We include this distinction, resulting in a total of $8\times 3 \times 10 = 240$ \fretish templates. Again, here we disregard the parameter $k$. Since all requirements involve a response and a component, these are not to be taken into consideration for the different templates. 
To sum up, \fretish consists of the following elements.

\begin{table}[h]
\centering
\renewcommand{\arraystretch}{1.2}
\begin{tabular}{p{0.45\linewidth}|p{0.51\linewidth}}
\multicolumn{1}{c}{\textbf{Mandatory Fields}} & 
\multicolumn{1}{c}{\textbf{Optional Fields}} \\ \hline

\begin{tabular}[t]{@{}l@{}}
\textbf{Component} \\
To which the requirement applies.
\end{tabular}
&
\begin{tabular}[t]{@{}l@{}}
\textbf{Timing} \\
When \textit{response} should occur.
\end{tabular}
\\[8pt]

\begin{tabular}[t]{@{}l@{}}
\textbf{Shall} \\
Verbal phrase indicating the\\  obligation.
\end{tabular}
&
\begin{tabular}[t]{@{}l@{}}
\textbf{Conditions} \\
Applicability context: \emph{void} \\ (unconditional), \emph{trigger} (activates \\on event), \emph{continual} (holds while true).
\end{tabular}
\\[8pt]

\begin{tabular}[t]{@{}l@{}}
\textbf{Response} \\
Required response behavior.
\end{tabular}
&
\begin{tabular}[t]{@{}l@{}}
\textbf{Scope} \\
Temporal intervals related to when the \\ response is required, built from a \textit{mode} \\(boolean) and operators.
\end{tabular}

\end{tabular}
\end{table}

\fretish is described in \cite{Giannakopoulou:2021:AutomatedFormalizationOfSNLReqs} as a Structured Natural Language (SNL). Sometimes, \fretish is also mentioned in the context of Controlled Natural Languages (CNL). The difference between both is a bit subtle, and the two are sometimes used as synonyms (\cite{mueckstein:1985:CNL}). As a rule of thumb, CNLs are used for the direction from humans to computers whereas, SNLs are used for the direction from computers to humans (see \cite{Gao:2019:controlled, Kuhn:2014:SurveyCNLs}). In this paper we are mainly concerned with the direction from human to computer whence we speak of Controlled Natural Language. In \cite{Giannakopoulou:2021:AutomatedFormalizationOfSNLReqs} an emphasis was on the translation of semantic templates to natural language whence they speak of Structured Natural Language. In a sense, \fretish can be conceived of as playing both roles (human to computer and vice-versa) so that the distinction between SNL and CNL is not that essential.

\subsection{The FRET tool}

\begin{figure}[h]
    \centering
    \includegraphics[width=0.8\linewidth]{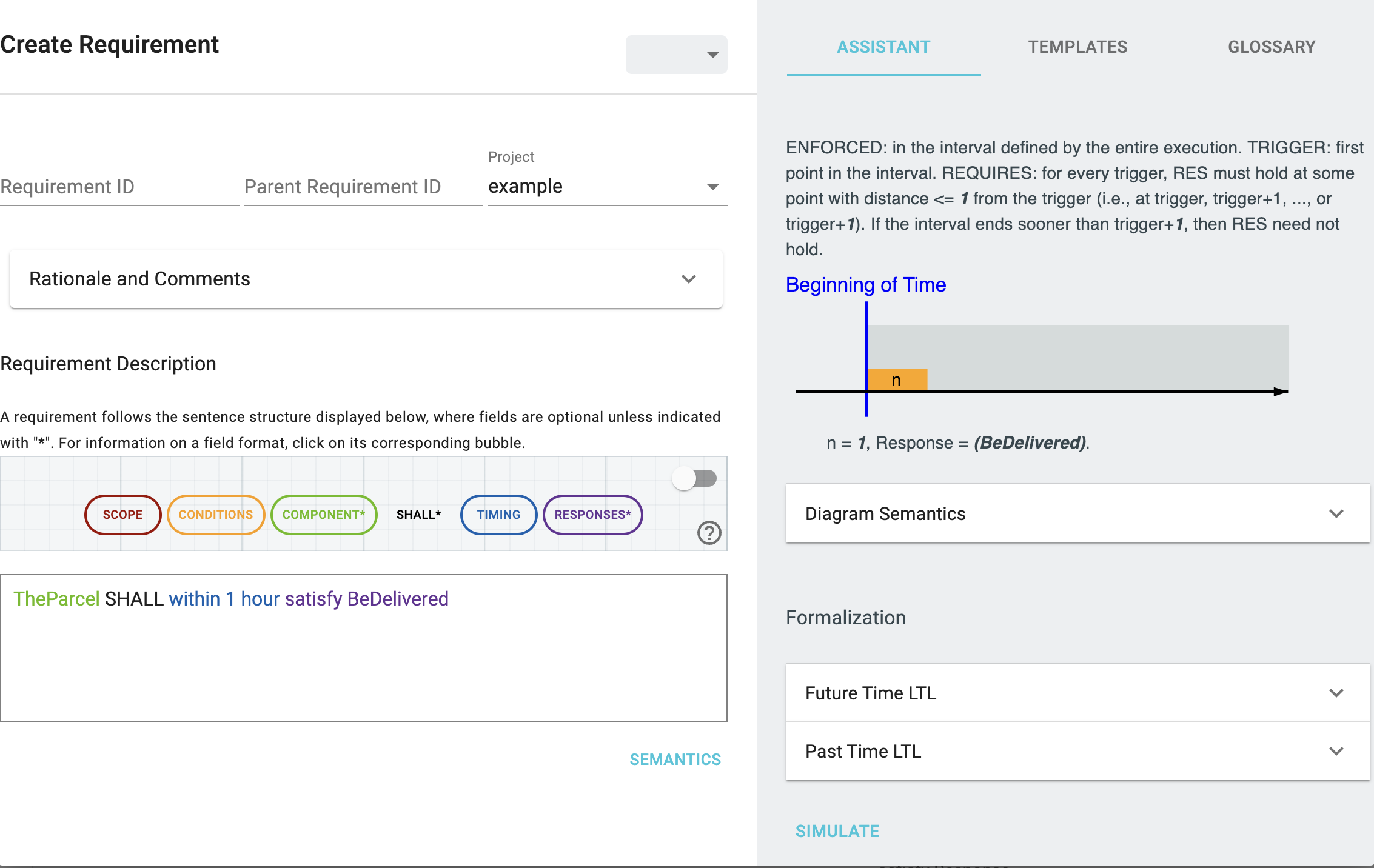}
    \caption{FRET user interface.}
\label{fig:ExampleFRETInterface}
\end{figure}

Figure \ref{fig:ExampleFRETInterface} illustrates an example on the FRET requirements elicitation interface. The natural language requirement ``The parcel shall be delivered within one hour" can be rephrased in \fretish as {\fret{}{}{TheParcel}{within 1 hour}{satisfy BeDelivered}}. 

Once the requirement is specified, the semantics pane provides multiple complementary representations to support analysis. In particular, it presents a structured textual interpretation of the \fretish statement, a semantic timeline diagram (sometimes called railroad diagram) that visually depicts when the requirement is enforced and how the response relates to triggering events over time, and formal translations into both future-time (finite trace and infinite trace) and past-time Metric Linear Temporal Logic (MTL). Together, these elements help users understand the temporal constraints, verify the correctness, and reason about the behavior of the requirement within the system execution.

In this paper, we focus on the translation functionality of FRET. However, FRET contains many more functionalities like an interactive simulator to help users create scenarios on which they can evaluate the requirements. We cite from the official git release \cite{fret-github}:
\begin{quote}
    FRET supports hierarchical requirement definition, enabling users to organize requirements based on different levels. It also offers export capabilities in various formats, facilitating integration with external formal analysis tools. FRET provides automated, requirement-based test case generation, incorporating a coverage metric to assess the adequacy of generated tests. Furthermore, FRET includes built-in consistency checking mechanisms to detect conflicts between requirements and thus, it helps to ensure that requirements are complete and correct.
\end{quote}

\section{Translations from \fretish into fragments of MTL}\label{section:Translations}

In this section, we will first comment on how \Fret translates \fretish into various fragments of MTL. Next, we will outline alternative translations that stick closer to the technical description of the semantic behavior provided by the FRET interface.

\subsection{From \fretish to MTL in FRET: the \NASA translation}

The first release of FRET is described at a very high level in \cite{GiannakopoulouEtAl:2020:FormalRE} where the design of \fretish is mentioned to incorporate features of various other specification languages like, the Specification Pattern System (SPS) \cite{DwyerEtAl1999SPS} and the Easy Approach to Requirements Syntax (EARS) \cite{Mavin:2012:EARS}. In \cite{GiannakopoulouEtAl:2020:GenerationFormalReqsFromSNL} and its journal extension \cite{Giannakopoulou:2021:AutomatedFormalizationOfSNLReqs} it is described how \fretish taps into the vast field of Interval Semantics \cite{Moszkowski:1982:IntervalLogic, HalpernEtAl:1991:TimeIntervalLogics, BruckerEtAl:2024:formallyVerifiedIntervalArithmetic} and in particular Real Time Graphical Interval Logic \cite{MoserEtAl:1996:RTGIL}.

When FRET translates a \fretish template to fMTL as an intermediate step, the \fretish template is translated into the Structured Assertion Language for Temporal logic (SALT ) \cite{BauerEtAl:2006:Salt, BauerEtAl:2011:SALT}. The semantics of \fretish templates is defined in terms of RTGIL templates and can be consulted in detail in \cite{GiannakopoulouEtAl:2020:FormalRE, Giannakopoulou:2021:AutomatedFormalizationOfSNLReqs, ConradEtAl:2022:CompositionalProofFrameworkFretish}. The NASA translations of \fretish requirements to the various LTL fragments follow the logical structure of the RTGIL semantics contrary to LTL or MTL semantics; we quote from \cite{Giannakopoulou:2021:AutomatedFormalizationOfSNLReqs}: ``At a high level, our algorithms view a trace as a collection of disjoint intervals".

There are three translations provided by the NASA tool FRET: fMTL on finite traces, fMTL on infinite traces and, pMTL which are understood to work on finite data traces. The  logic pMTL is often used in real-time run-time verification \cite{BartocciFalcone:2018:LecturesNotesRuntimeVerification}, where real-time data records are checked against their specifications. Various tools, like Lustre-based analysis tools as Kind2 \cite{ChampionEtAl:2016:Kind2ModelChecker} and JKind \cite{GacekEtAl:2018:JkindModelChecker} only support past time. However, requirements are typically formulated using future time constructs such as  \textit{within the next 10 seconds}, \textit{during 2 minutes}, \textit{until} $\sf stop$, etc. Thus, rephrasing these requirements exclusively in terms of past time logic seems artificial and this practice is certainly not in line with (N1) from above. It seems that past time evaluation of requirements begs the use of full MTL (both past and future time).

\subsection{An alternative translation from \fretish to LTL: The \FV translation}

When we started studying \fretish we found it difficult to understand the MTL formalisations and to reconcile them to the corresponding technical description provided by FRET. The technical description always had the form ``If something scope and condition related happens then some timed response will occur" (our core implication $\Phi$ below). We then started out from scratch to design MTL formulas with not only exactly this semantical behavior but also exactly this semantical structure (whence yielding coherency).

We define the \FV translation as a compositional mapping from \fretish requirements to MTL formulas.
Each requirement is decomposed into four canonical components: \textit{scope}, \textit{condition}, \textit{timing}, and \textit{response},
as in FRET~\cite{mavridou2023formal}.
Intuitively, a requirement specifies:
(i) a set of intervals where it is enforced (scope),
(ii) a set of triggering points within those intervals (condition), and
(iii) a constrained system behavior (timing and response).
Where \cite{ConradEtAl:2022:CompositionalProofFrameworkFretish} only included trigger conditions, we decided to include both continual and trigger conditions. 
Below, we map combinations of FRET templates that we encode by $\langle s, c, t \rangle$ out of the 180 possible templates described in Section \ref{section:FRET}.

\begin{figure}[t]
\centering

\begin{minipage}[t]{0.37\textwidth}
\vspace{0pt}
\centering
\resizebox{\linewidth}{!}{%
\small
\begin{tabular}{l l}
\textbf{Scope} & \textbf{LTL characterization} \\
\midrule

Global & $\top$ \\
In \scope & $\scope$ \\
Not in \scope & $\neg \ \scope$ \\
Only in \scope & $\neg \ \scope$ \\
Before \scope & $\Hist \,\neg \ \scope$ \\
Only before \scope & $\On \scope$ \\
After \scope & $\On (\neg \ \scope \land \Y\,\scope)$ \\
Only after \scope & $\Hist \,(\Y \scope \rightarrow \scope)$ \\

\bottomrule
\end{tabular}
}
\captionof{table}{LTL characterization of Scope operators.}
\label{tab:scopes}
\end{minipage}
\hfill
\begin{minipage}[t]{0.55\textwidth}
\vspace{0pt}
\centering

\resizebox{\linewidth}{!}{%
\begin{tikzpicture}

\draw [thick, ->] (0,0) -- (20,0);

\node[circle,fill=black,inner sep=0pt,minimum size=3pt,label=below:{\footnotesize$ftp$}] at (0,0) {};
\node[circle,fill=black,inner sep=0pt,minimum size=3pt,label=below:{\footnotesize$t_1{-}1$}] at (3,0) {};
\node[circle,fill=black,inner sep=0pt,minimum size=3pt,label=below:{\footnotesize$t_1$}] at (4,0) {};
\node[circle,fill=black,inner sep=0pt,minimum size=3pt,label=below:{\footnotesize$t_2$}] at (7,0) {};
\node[circle,fill=black,inner sep=0pt,minimum size=3pt,label=below:{\footnotesize$t_2{+}1$}] at (8,0) {};
\node[circle,fill=black,inner sep=0pt,minimum size=3pt,label=below:{\footnotesize$t_3{-}1$}] at (11,0) {};
\node[circle,fill=black,inner sep=0pt,minimum size=3pt,label=below:{\footnotesize$t_3$}] at (12,0) {};
\node[circle,fill=black,inner sep=0pt,minimum size=3pt,label=below:{\footnotesize$t_4$}] at (15,0) {};
\node[circle,fill=black,inner sep=0pt,minimum size=3pt,label=below:{\footnotesize$t_4{-}1$}] at (16,0) {};
\node[circle,fill=black,inner sep=0pt,minimum size=3pt,label=below:{\footnotesize$n$}] at (19,0) {};

\foreach \x in {0,3,4,7,8,11,12,15,16,19} {
  \draw [thick, gray, dashed] (\x,0) -- (\x,10.5);
}

\node[rectangle, draw=steelblue, fill=steelblue, text=white, minimum width=3cm, minimum height=1cm] at (5.5,.5) {MODE};
\node[rectangle, draw=steelblue, fill=steelblue, text=white, minimum width=3cm, minimum height=1cm] at (13.5,.5) {MODE};

\node[rectangle, draw=steelblue, fill=lightgray, text=black, minimum width=3cm, minimum height=1cm] at (5.5,2) {IN};
\node[rectangle, draw=steelblue, fill=lightgray, text=black, minimum width=3cm, minimum height=1cm] at (13.5,2) {IN};

\node[rectangle, draw=steelblue, fill=lightgray, text=black, minimum width=3cm, minimum height=1cm] at (1.5,3.5) {BEFORE};

\node[rectangle, draw=steelblue, fill=lightgray, text=black, minimum width=11cm, minimum height=1cm] at (13.5,4) {AFTER};

\node[rectangle, draw=steelblue, fill=lightgray, text=black, minimum width=3cm, minimum height=1cm] at (1.5,5.5) {NOTIN};
\node[rectangle, draw=steelblue, fill=lightgray, text=black, minimum width=3cm, minimum height=1cm] at (9.5,5.5) {NOTIN};
\node[rectangle, draw=steelblue, fill=lightgray, text=black, minimum width=3cm, minimum height=1cm] at (17.5,5.5) {NOTIN};

\node[rectangle, draw=steelblue, fill=lightgray, text=black, minimum width=3cm, minimum height=1cm] at (1.5,7) {ONLYIN};
\node[rectangle, draw=steelblue, fill=lightgray, text=black, minimum width=3cm, minimum height=1cm] at (9.5,7) {ONLYIN};
\node[rectangle, draw=steelblue, fill=lightgray, text=black, minimum width=3cm, minimum height=1cm] at (17.5,7) {ONLYIN};

\node[rectangle, draw=steelblue, fill=lightgray, text=black, minimum width=15cm, minimum height=1cm] at (11.5,8.5) {ONLYBEFORE};

\node[rectangle, draw=steelblue, fill=lightgray, text=black, minimum width=7cm, minimum height=1cm] at (3.5,10) {ONLYAFTER};

\end{tikzpicture}
}

\captionof{figure}{Semantic behavior of Scope kinds.}
\label{fig:scope}
\end{minipage}

\end{figure}

We 
define
two kinds of intermediate functions: time-independent functions and time-dependent functions. In the first group, we define $\mathsf{mtl\_of\_scope}(s)$, which is encoded in Table~\ref{tab:scopes}, and the formula $\mathsf{triggers}(s,c)$, whose definition is as follows:
\[
\mathsf{triggers}(s,c) :=
\begin{cases}
\mathsf{mtl\_of\_scope}(s) \land \condition
& \text{(continual condition)} \\[0.6em]

\begin{aligned}[t]
&(\mathsf{mtl\_of\_scope}(s) \land \condition) \\
& \ \ \land Z \neg(\mathsf{mtl\_of\_scope}(s) \land \condition)
\end{aligned}
& \text{(trigger condition)}
\end{cases}
\]

So, when $c$ is continual, we go with the upper line, and we go with the lower line if $c$ is a trigger condition. If $c$ is absent, we go with the lower line taking $\condition = \top$. We could have simplified that expression but for the sake of uniformity decided not to do so. 

These components are combined into a core implication:
\[
\Phi(s,c,t,\sigma) := 
\mathsf{triggers}(s,c) \rightarrow 
\mathsf{timed\_response}_{\sigma}(t,s)
\]
where $\sigma \in \{\mathsf{past}, \mathsf{fin}, \mathsf{inf}\}$.\\
\medskip

To support different execution models, the formula $\Phi$ is lifted to a global temporal specification:

\[
\begin{aligned}
\textsc{Past-time:} \quad & \Hist(\Phi) \\
\textsc{Future (infinite):} \quad & \G(\Phi) \\
\textsc{Future (finite):} \quad & (\last) \R \Phi
\end{aligned}
\]

Thus, the final translation is:
\[
\mathsf{mtl\_of\_fretish}(f) :=
\begin{cases}
\G(\Phi) & \text{(infinite traces)} \\
(\last) \R \Phi & \text{(finite traces)} \\
\Hist(\Phi) & \text{(past-time)}
\end{cases}
\]

According to the authors of \cite{Giannakopoulou:2021:AutomatedFormalizationOfSNLReqs}, one of the improvements over \cite{GiannakopoulouEtAl:2020:GenerationFormalReqsFromSNL} is an improvement in the formalisation so that it ``highlight[s] the high-level structure of our approach, and the commonalities between fmLTL and pmLTL formalisations."
We argue that the \FV translation improves even further on these communalities.

FRET supports multiple semantic interpretations of time, allowing requirements to be evaluated under past-time, future-time (infinite traces), and future-time (finite traces) semantics. As a consequence, the interpretation of the \textit{timing} component is not uniform, but depends on the chosen trace model.
We therefore refine the timing semantics into three variants:
\[
\mathsf{timed\_response}_{\mathsf{past}}, \quad
\mathsf{timed\_response}_{\mathsf{fin}}, \quad
\mathsf{timed\_response}_{\mathsf{inf}},
\]
corresponding respectively to past-time, finite-trace future-time, and infinite-trace future-time semantics, encoding the different columns in Table~\ref{tab:timing_semantics}.

Table~\ref{tab:timing_semantics} summarizes the semantics of each timing operator.
Rows correspond to timing patterns and scope variants (\emph{regular} vs.\ \emph{only}),
while columns correspond to different trace semantics. To define them, we introduce some common patterns.

We identify recurring subformula patterns, such as \textit{scope exit conditions} and \textit{scope-restricted Until operators}. Let $\phi_s$ denote the scope condition. We define a derived operator:
\[
p \U_{\phi_s} q := p \U (q \land \phi_s).
\]

This operator enforces that $q$ must occur while the scope remains active.
In addition, we introduce two auxiliary predicates that capture structural properties of the scope:

\begin{itemize}
    \item \textbf{Scope condition} $\phi_s$: this is simply and abbreviation of $\mathsf{mtl\_of\_scope}$.
    
    \item \textbf{Scope exit} $\mathit{exit}_s$: a predicate identifying the last instant in which the scope holds. In future-time semantics, this can be characterized as $
    \mathit{exit}_s := (\phi_s \land \X \neg \phi_s) \lor \last$.    
\end{itemize}

Intuitively, $\mathit{exit}_s$ holds either when the system is about to leave the scope in the next time step, or when the end of the trace is reached. This notion is particularly relevant in finite-trace semantics, where obligations must be evaluated with respect to the last state.

These constructs appear systematically in the formalisation of timing patterns. In particular:

\begin{itemize}
    \item Scope-restricted Until ($\U_{\phi_s}$) ensures that eventualities are fulfilled before leaving the scope.
    
    \item The exit condition $\mathit{exit}_s$ is used to encode safety-like constraints, typically through the Release operator, ensuring that properties hold up to the boundary of the scope.
\end{itemize}

As a result, many timing patterns can be expressed using a small set of recurring logical forms involving $\phi_s$, $\mathit{exit}_s$, and $\U_{\phi_s}$.

Since we started our translation from scratch, we found that we sometimes would have made other choices than the design choices of the \NASA translation. Most importantly, our understanding of how scopes work that involve the word ``only'' turned out to be different from the NASA implementation and, to our understanding, to the NASA description.

From \cite{Giannakopoulou:2021:AutomatedFormalizationOfSNLReqs} we quote
    \begin{quote}
        It is sometimes necessary to specify that a requirement is enforced only in some time frame, meaning that it should not be satisfied outside of that frame. For this, the scopes only after, only before, and only in are provided;\\
        \ldots\\
        Note that only* scopes, mandate that a requirement not hold outside of their corresponding scope.
    \end{quote}
Our interpretation of this would have been to work with 
\[
\Phi'(s,c,t,\sigma) := 
\neg \mathsf{triggers}'(s,c) \rightarrow 
\neg \mathsf{timed\_response}_{\sigma}(t,s)
\]
instead of $\Phi(s,c,t,\sigma) := 
\mathsf{triggers}(s,c) \rightarrow 
\mathsf{timed\_response}_{\sigma}(t,s)
$. Here, $\mathsf{triggers}'(s,c)$ would be obtained by substituting $\neg \mathsf{mtl\_of\_scope}(s)$ for $\mathsf{mtl\_of\_scope}(s)$ in $\mathsf{triggers}(s,c)$. It turns out that NASA makes another choice here which is harder to explain and we refer the reader to \cite{ConradEtAl:2022:CompositionalProofFrameworkFretish} for details. We decided to keep our general structure and make minor adjustments on the only case so to remain equivalent with the \NASA translation. We acted similarly if we would have given a different interpretation to a timed response. We comment on that in the next section.

\begin{table}[H]
\footnotesize
\centering
\renewcommand{\arraystretch}{1.15}
\begin{tabular}{l c p{4.2cm}  p{4.2cm}  p{4.2cm}}
\toprule
\textbf{Timing} & \textbf{Scope} & \textbf{Future Inf.} & \textbf{Future Fin.} & \textbf{Past} \\
\midrule

\rowcolor{blockgray}
\multicolumn{5}{l}{\textbf{Unbounded}} \\
\addlinespace[2pt]

\multirow{2}{*}{immediately}
& \NotOnly 
& $\response$
& $\response$
& $\response$ \\

& \cellcolor{rowgray}\Only 
& \cellcolor{rowgray}$\neg \response$
& \cellcolor{rowgray}$\neg \response$
& \cellcolor{rowgray}$\neg \response$ \\

\multirow{2}{*}{eventually}
& \NotOnly 
& $\phi_s \Us \response$
& $(\neg \last \land \phi_s)\Us \response$
& \sameInf \\

& \cellcolor{rowgray}\Only 
& \cellcolor{rowgray}$\neg (\phi_s \Us \response)$
& \cellcolor{rowgray}$\neg ((\neg \last \land \phi_s)\Us \response)$
& \cellcolor{rowgray}$\exitS \R \neg \response$ \\

\multirow{2}{*}{next}
& \NotOnly 
& $\neg \exitS \rightarrow \X \response$
& \sameInf
& \sameInf \\

& \cellcolor{rowgray}\Only 
& \cellcolor{rowgray}$\neg \exitS \rightarrow \X \neg \response$
& \cellcolor{rowgray}\sameInf
& \cellcolor{rowgray}\sameInf \\

\multirow{2}{*}{always}
& \NotOnly 
& $\exitS \R \response$
& \sameInf
& \sameInf \\

& \cellcolor{rowgray}\Only 
& \cellcolor{rowgray}$\neg (\exitS \R \response)$
& \cellcolor{rowgray}\sameInf
& \cellcolor{rowgray}$(\neg \last \land \phi_s)\Us \neg \response$ \\

\multirow{2}{*}{never}
& \NotOnly 
& $\exitS \R \neg \response$
& \sameInf
& \sameInf \\

& \cellcolor{rowgray}\Only 
& \cellcolor{rowgray}$\neg (\exitS \R \neg \response)$
& \cellcolor{rowgray}\sameInf
& \cellcolor{rowgray}$(\neg \last \land \phi_s)\Us \response$ \\

\midrule
\rowcolor{blockgray}
\multicolumn{5}{l}{\textbf{Bounded}} \\
\addlinespace[3pt]

\multirow{2}{*}{within $[0,k]$}
& \NotOnly 
& $\Fb{0,k}\response \lor \exitWithin{k}$
& $\Fb{0,k}\response \lor \Fb{0,k-1}\exitS$
& \sameInf \\

& \cellcolor{rowgray}\Only 
& \cellcolor{rowgray}$\Gb{0,k}\neg \response \lor ((\neg \response \land \phi_s)\Ub{0,k}\neg \phi_s)$
& \cellcolor{rowgray}$\Gb{0,k}\neg \response \lor (\exitS \R \neg \response)$
& \cellcolor{rowgray}\sameInf \\

\multirow{2}{*}{for $[0,k]$}
& \NotOnly 
& $\Gb{0,k}\response \lor ((\response \land \phi_s)\Ub{0,k}\neg \phi_s)$
& $\Gb{0,k}\response \lor (\exitS \R \response)$
& \sameInf \\

& \cellcolor{rowgray}\Only 
& \cellcolor{rowgray}$\Fb{0,k}\neg \response \lor \exitWithin{k}$
& \cellcolor{rowgray}$\Fb{0,k}\neg \response \lor \Fb{0,k-1}\exitS$
& \cellcolor{rowgray}\sameInf \\

\multirow{2}{*}{after $[0,k]$}
& \NotOnly
& $(\Gb{0,k}\neg \response \land \X^{k+1}\response) \lor ((\neg \response \land \phi_s)\Ub{0,k+1}\neg \phi_s)$
& $(\Gb{0,k}\neg \response \lor (\exitS \R \neg \response)) \land (\Fb{0,k+1}\response \lor \Fb{0,k}\exitS)$
& $(\Gb{0,k}\neg \response \land (\neg \response \Ub{0,k+1}\response)) \lor ((\neg \response \land \phi_s)\Ub{0,k+1}\neg \phi_s)$ \\

& \cellcolor{rowgray}\Only 
& \cellcolor{rowgray}$(\Fb{0,k}\response \lor \Gb{0,k+1}\neg \response) \lor (\phi_s \Ub{0,k+1}\neg \phi_s)$
& \cellcolor{rowgray}$(\Fb{0,k}\response \lor \Fb{0,k-1}\exitS) \lor (\Gb{0,k+1}\neg \response \lor (\exitS \R \neg \response))$
& \cellcolor{rowgray}\sameInf \\

\midrule

\rowcolor{blockgray}
\multicolumn{5}{l}{\textbf{Condition-based}} \\
\addlinespace[2pt]

\multirow{2}{*}{until $\stopCondition$}
& \NotOnly 
& $\G \response \lor (\response \U (\stopCondition \lor (\neg \phi_s \land \Y \phi_s)))$
& $(\response \land \phi_s)\U ((\response \land \last) \lor \stopCondition \lor \neg \phi_s)$
& \sameInf \\

& \cellcolor{rowgray}\Only 
& \cellcolor{rowgray}$(\neg \response \R \neg \stopCondition) \lor (\exitS \R \neg \stopCondition)$
& \cellcolor{rowgray}$(\neg \response \lor \exitS)\R \neg \stopCondition$
& \cellcolor{rowgray}\sameInf \\

\multirow{2}{*}{before $\stopCondition$}
& \NotOnly 
& $(\response \R \neg \stopCondition) \lor (\exitS \R \neg \stopCondition)$
& $(\response \lor \exitS)\R \neg \stopCondition$
& \sameInf \\

& \cellcolor{rowgray}\Only 
& \cellcolor{rowgray}$\G \neg \response \lor (\neg \response \U (\stopCondition \lor (\neg \phi_s \land \Y \phi_s)))$
& \cellcolor{rowgray}$((\stopCondition \lor \exitS)\R (\neg \response \lor \stopCondition)) \lor (\neg \response \land \neg \phi_s \land \X \neg \phi_s) \lor (\exitS \R \neg \response)$
& \cellcolor{rowgray}\sameInf \\

\bottomrule
\end{tabular}
\caption{
Comparison of timing semantics across infinite, finite, and past interpretations.
Formulas are parameterized by the scope condition $\phi_s$, and are classified according to \emph{regular} and \emph{only} scopes.}
\label{tab:timing_semantics}
\end{table}

\medskip

\section{Implementation and comparison}\label{section:Comparison}

In this section we compare the two translations and describe how we did so.

\subsection{Our production pipeline}

We considered all 240 \fretish templates described in Section \ref{section:FRET} and generated the \NASA translation using the FRET tool and engine. We wrote a Python script to generate the corresponding \FV translations into \texttt{nuXmv}-compatible syntax~\cite{cav15-nuxmv}, enabling integration into verification and testing pipelines.

The semantic correctness of the \FV translation has been validated through systematic model checking in \texttt{nuXmv}, covering a representative set of requirement patterns. In particular, for each combination of scope, condition, timing\footnote{For bounded timings, we fixed the parameter to $k = 3$ and conducted some experiments to see that any other choice would yield similar results.}, and response, 
we verified equivalence by checking that the formula 
\[
\G (\NASA(s, c, t, r) \leftrightarrow \FV(s,c,t,r))
\]
is a tautology using model checking.
For finite traces, we encoded the end of the trace explicitly using a dedicated propositional variable $\last$, which becomes true at the final point of the trace and remains true thereafter.\footnote{This standard encoding is commonly used and allows finite traces to be represented within standard infinite-trace semantics.}
Finally, we performed a comparative analysis of the two translations by measuring the structural complexity of the resulting formulas, including size, temporal depth, and number of propositional occurrences.

\subsection{Quantitative comparison}

The quantitative analysis is presented at two levels: a global evaluation and a set of representative case studies.
For the global analysis, we considered all 240 possible requirements of the form
\fret{Scope}{Condition}{MyComponent}{timing}{response},
and computed aggregate metrics including the average formula size, temporal depth, and number of propositional occurrences. These results are grouped by timing pattern and presented in Table~\ref{tab:metrics_fin}.

\begin{table}[H]
\footnotesize
\centering
\begin{tabular}{lcccc}
\toprule
Timing & Trans & av. Len & av. Depth & av. Props Used \\
\midrule

\multirow{2}{*}{immediately}
& \sf{FV}   & 12.96 & 2.42 & 3.88 \\
& \cellcolor{rowgray}\sf{FRET} & \cellcolor{rowgray}42.71 & \cellcolor{rowgray}2.46 & \cellcolor{rowgray}9.67 \\

\multirow{2}{*}{eventually}
& \sf{FV} & 23.46 & 2.75 & 6.12 \\
& \cellcolor{rowgray}FRET & \cellcolor{rowgray}58.17 & \cellcolor{rowgray}3.75 & \cellcolor{rowgray}12.04 \\

\multirow{2}{*}{at the next timepoint}
& \sf{FV}   & 26.46 & 3.75 & 6.12 \\
& \cellcolor{rowgray}\sf{FRET} & \cellcolor{rowgray}76.71 & \cellcolor{rowgray}3.25 & \cellcolor{rowgray}15.00 \\

\multirow{2}{*}{always}
& \sf{FV} & 25.46 & 4.75 & 6.12 \\
& \cellcolor{rowgray}\sf{FRET} & \cellcolor{rowgray}56.38 & \cellcolor{rowgray}3.62 & \cellcolor{rowgray}11.75 \\

\multirow{2}{*}{never}
& \sf{FV}   & 26.46 & 4.75 & 6.12 \\
& \cellcolor{rowgray}\sf{FRET} & \cellcolor{rowgray}56.54 & \cellcolor{rowgray}3.62 & \cellcolor{rowgray}11.75 \\

\multirow{2}{*}{within 3 seconds}
& \sf{FV}   & 27.21 & 3.75 & 6.50 \\
& \cellcolor{rowgray}\sf{FRET} & \cellcolor{rowgray}64.38 & \cellcolor{rowgray}3.88 & \cellcolor{rowgray}13.17 \\

\multirow{2}{*}{for 3 seconds}
& \sf{FV}   & 27.71 & 4.38 & 6.75 \\
& \cellcolor{rowgray}\sf{FRET} & \cellcolor{rowgray}63.88 & \cellcolor{rowgray}3.88 & \cellcolor{rowgray}13.50 \\

\multirow{2}{*}{after 3 seconds}
& \sf{FV}   & 44.58 & 3.75 & 10.38 \\
& \cellcolor{rowgray}\sf{FRET} & \cellcolor{rowgray}91.88 & \cellcolor{rowgray}3.88 & \cellcolor{rowgray}19.00 \\

\multirow{2}{*}{until StopCondition}
& \sf{FV}   & 27.46 & 3.12 & 7.75 \\
& \cellcolor{rowgray}\sf{FRET} & \cellcolor{rowgray}100.38 & \cellcolor{rowgray}3.88 & \cellcolor{rowgray}22.33 \\

\multirow{2}{*}{before StopCondition}
& \sf{FV}   & 37.96 & 3.75 & 10.25 \\
& \cellcolor{rowgray}\sf{FRET} & \cellcolor{rowgray}96.71 & \cellcolor{rowgray}3.88 & \cellcolor{rowgray}21.00 \\

\bottomrule
\end{tabular}
\caption{Average formula metrics for \FV and \NASA under finite semantics}
\label{tab:metrics_fin}
\end{table}

Complementing the analysis at the level of averages, Table~\ref{tab:big_req} presents detailed results for a selection of representative requirements.

\begin{table}[H]
\footnotesize
\centering
\renewcommand{\arraystretch}{1.2}

\begin{tabular}{p{7cm} l r r r r}
\toprule
Requirement & Tool & Temp Ops & Props & TDepth & Size \\
\midrule

\multirow{2}{=}{\fret{In Scope}{upon Condition}{MyComponent}{until StopCondition}{satisfy Response}}
& \sf{FV}   & 3  & 9  & 2 & 24 \\
& \cellcolor{rowgray}\sf{FRET} & \cellcolor{rowgray}35 & \cellcolor{rowgray}65 & \cellcolor{rowgray}6 & \cellcolor{rowgray}303 \\

\multirow{2}{=}{\fret{Not in Scope}{upon Condition}{MyComponent}{until StopCondition}{satisfy Response}}
& \sf{FV}   & 3  & 9  & 2 & 28 \\
& \cellcolor{rowgray}\sf{FRET} & \cellcolor{rowgray}35 & \cellcolor{rowgray}65 & \cellcolor{rowgray}6 & \cellcolor{rowgray}304 \\

\multirow{2}{=}{\fret{Only in Scope}{upon Condition}{MyComponent}{after 3 seconds}{satisfy Response}}
& \sf{FV}   & 8  & 11 & 3 & 44 \\
& \cellcolor{rowgray}\sf{FRET} & \cellcolor{rowgray}39 & \cellcolor{rowgray}49 & \cellcolor{rowgray}6 & \cellcolor{rowgray}244 \\

\multirow{2}{=}{\fret{Only in Scope}{upon Condition}{MyComponent}{before StopCondition}{satisfy Response}}
& \sf{FV}   & 7  & 15 & 3 & 53 \\
& \cellcolor{rowgray}\sf{FRET} & \cellcolor{rowgray}35 & \cellcolor{rowgray}65 & \cellcolor{rowgray}6 & \cellcolor{rowgray}316 \\

\bottomrule
\end{tabular}
\caption{Comparison of \textsc{Fv} and \Fret translations for representative \fretish requirements (finite semantics).}
\label{tab:big_req}
\end{table}

\begin{figure}[H]
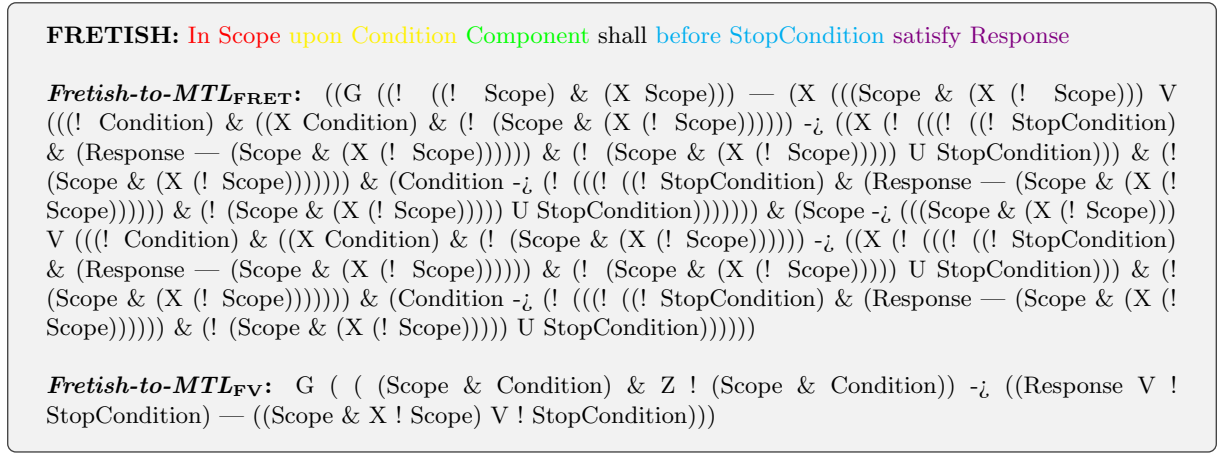

\begin{center}
\begin{tcolorbox}[colback=gray!10, boxrule=0.5pt, width=\textwidth]
\footnotesize
\textbf{FRETISH:} \fret{In Scope}{upon Condition}{Component}{before StopCondition}{satisfy Response}\\
\ \\
\textbf{\NASA:} ((G ((! ((! Scope) \& (X Scope))) | (X (((Scope \& (X (! Scope))) V (((! Condition) \& ((X Condition) \& (! (Scope \& (X (! Scope)))))) -> ((X (! (((! ((! StopCondition) \& (Response | (Scope \& (X (! Scope)))))) \& (! (Scope \& (X (! Scope))))) U StopCondition))) \& (! (Scope \& (X (! Scope))))))) \& (Condition -> (! (((! ((! StopCondition) \& (Response | (Scope \& (X (! Scope)))))) \& (! (Scope \& (X (! Scope))))) U StopCondition))))))) \& (Scope -> (((Scope \& (X (! Scope))) V (((! Condition) \& ((X Condition) \& (! (Scope \& (X (! Scope)))))) -> ((X (! (((! ((! StopCondition) \& (Response | (Scope \& (X (! Scope)))))) \& (! (Scope \& (X (! Scope))))) U StopCondition))) \& (! (Scope \& (X (! Scope))))))) \& (Condition -> (! (((! ((! StopCondition) \& (Response | (Scope \& (X (! Scope)))))) \& (! (Scope \& (X (! Scope))))) U StopCondition))))))\\ 
\ \\
\textbf{\FV:} G ( ( (Scope \& Condition) \& Z ! (Scope \& Condition)) -> ((Response V ! StopCondition) | ((Scope \& X ! Scope) V ! StopCondition)))

\end{tcolorbox}
\caption{A particular \fretish formula and their respective MTL translations on infinite traces.}
\label{Figure:TranslationsParticularFormula}
\end{center}
\end{figure}

The representative requirements from Table \ref{tab:big_req} have been chosen to highlight scenarios where the compositional interaction between scope, condition, and timing becomes non-trivial. In particular, we focus on condition-based patterns such as \textit{until StopCondition}, as well as scope-restricted variants involving \emph{Only}, where the structural complexity of the translations is most pronounced.
It may also be instructive to see a particular \fretish requirement and their particular corresponding translations. This is exhibited in Figure 
\ref{Figure:TranslationsParticularFormula}. 

The \NASA translation consistently produces substantially larger formulas than the corresponding \FV translation which is often an order of magnitude higher in terms of total size and number of propositional occurrences. The effect is especially pronounced in patterns where multiple semantic dimensions interact simultaneously: for instance, when a timing constraint (e.g., \textit{until}) must be evaluated relative to scope boundaries and triggering conditions. In such cases, the \Fret translation expands into deeply nested structures that explicitly encode all boundary conditions and edge cases.

Interestingly, this increase in size is not always accompanied by a proportional increase in temporal depth. As shown in the tables \ref{tab:FinSemDepthPast} and \ref{tab:InfSemDepth} below, the temporal nesting remains relatively moderate, suggesting that the added complexity is primarily due to propositional duplication and structural redundancy rather than deeper temporal reasoning. In other words, the formulas become wider rather than deeper.

This observation has several implications. First, from a verification perspective, larger formulas may negatively impact model checking performance, as they increase the size of the symbolic representation and the cost of simplification. Second, from a human-understandability standpoint, the lack of a coherent and modular structure makes the formulas more difficult to inspect, debug, and relate back to the original requirement.

We observe that the current FRET translation increases the number of occurrences of propositional
variables in a formula compared to our proposed translation. This is particularly relevant for
test-case generation \cite{PecheurEtAl:2009:AnalysisReqBasedTesting}, \cite{KatisEtAl:2025:StreamlinedReqBasedTesting}, as trap formulas are generated per occurrence of propositional variables.
Consequently, the \FV translation yields a smaller number of trap formulas, which can lead to more
compact and efficient test suites.

Moreover, realizability checking as presented in 
\cite{KatisEtAl:2022:RealizabilityCheckingInFRET, KooiMavridou:2019:RealizabilityCheckingInFRET, GiannakopoulouEtAl:2021:RealizabilityWithinFRET} could not be reproduced in our setting,
as the required tool support was not available. To the best of our knowledge, neither nuXmv
nor Kind 2 provide support for realizability checking of specifications expressed in full MTL, in
particular those involving future-time operators.

\begin{table}[H]
\centering

\begin{minipage}{0.48\textwidth}
\centering
\begin{table}[H]
\footnotesize
\centering
\begin{tabular}{lcccc}
\toprule
Timing & Trans & Len & Depth & Props \\
\midrule
\multirow{2}{*}{immediately}
& \sf{FV} & 11.96 & 2.42 & 3.88 \\
& \cellcolor{rowgray}\sf{FRET} & \cellcolor{rowgray}29.08 & \cellcolor{rowgray}2.46 & \cellcolor{rowgray}9.67 \\ 
\multirow{2}{*}{eventually}
& \sf{FV} & 19.46 & 2.75 & 6.12 \\
& \cellcolor{rowgray}\sf{FRET} & \cellcolor{rowgray}37.54 & \cellcolor{rowgray}3.75 & \cellcolor{rowgray}12.04 \\ 
\multirow{2}{*}{at the next timepoint}
& \sf{FV} & 23.46 & 3.75 & 6.12 \\
& \cellcolor{rowgray}\sf{FRET} & \cellcolor{rowgray}48.42 & \cellcolor{rowgray}3.25 & \cellcolor{rowgray}15.0 \\ 
\multirow{2}{*}{always}
& \sf{FV} & 22.46 & 4.75 & 6.12 \\
& \cellcolor{rowgray}\sf{FRET} & \cellcolor{rowgray}36.42 & \cellcolor{rowgray}3.62 & \cellcolor{rowgray}11.75 \\ 
\multirow{2}{*}{never}
& \sf{FV} & 23.46 & 4.75 & 6.12 \\
& \cellcolor{rowgray}\sf{FRET} & \cellcolor{rowgray}36.58 & \cellcolor{rowgray}3.62 & \cellcolor{rowgray}11.75 \\ 
\multirow{2}{*}{within 3 seconds}
& \sf{FV} & 22.58 & 2.75 & 6.5 \\
& \cellcolor{rowgray}\sf{FRET} & \cellcolor{rowgray}42.75 & \cellcolor{rowgray}3.88 & \cellcolor{rowgray}13.17 \\ 
\multirow{2}{*}{for 3 seconds}
& \sf{FV} & 22.71 & 2.75 & 6.75 \\
& \cellcolor{rowgray}\sf{FRET} & \cellcolor{rowgray}42.25 & \cellcolor{rowgray}3.88 & \cellcolor{rowgray}13.5 \\ 
\multirow{2}{*}{after 3 seconds}
& \sf{FV} & 26.96 & 2.75 & 7.75 \\
& \cellcolor{rowgray}\sf{FRET} & \cellcolor{rowgray}62.25 & \cellcolor{rowgray}3.88 & \cellcolor{rowgray}19.0 \\ 
\multirow{2}{*}{until StopCondition}
& \sf{FV} & 26.83 & 3.75 & 8.12 \\
& \cellcolor{rowgray}\sf{FRET} & \cellcolor{rowgray}66.92 & \cellcolor{rowgray}3.88 & \cellcolor{rowgray}22.0 \\ 
\multirow{2}{*}{before StopCondition}
& \sf{FV} & 27.46 & 3.75 & 8.12 \\
& \cellcolor{rowgray}\sf{FRET} & \cellcolor{rowgray}66.25 & \cellcolor{rowgray}3.88 & \cellcolor{rowgray}21.0 \\ 
\bottomrule
\end{tabular}
\caption{Comparison of \Fret vs \textsc{Fv} (infinite semantics)}
\label{tab:InfSemDepth}
\end{table}
\end{minipage}
\hfill
\begin{minipage}{0.48\textwidth}
\centering
\begin{table}[H]
\footnotesize
\centering
\begin{tabular}{lcccc}
\toprule
Timing & Trans & Len & Depth & Props \\
\midrule
\multirow{2}{*}{immediately}
& \sf{FV} & 11.96 & 2.42 & 3.88 \\
& \cellcolor{rowgray}\sf{FRET} & \cellcolor{rowgray}43.58 & \cellcolor{rowgray}3.08 & \cellcolor{rowgray}13.42 \\ 
\multirow{2}{*}{eventually}
& \sf{FV} & 22.08 & 3.12 & 6.12 \\
& \cellcolor{rowgray}\sf{FRET} & \cellcolor{rowgray}51.71 & \cellcolor{rowgray}3.5 & \cellcolor{rowgray}15.21 \\ 
\multirow{2}{*}{at the next timepoint}
& \sf{FV} & 22.46 & 2.75 & 6.12 \\
& \cellcolor{rowgray}\sf{FRET} & \cellcolor{rowgray}67.04 & \cellcolor{rowgray}4.17 & \cellcolor{rowgray}19.58 \\ 
\multirow{2}{*}{always}
& \sf{FV} & 21.83 & 3.38 & 6.12 \\
& \cellcolor{rowgray}\sf{FRET} & \cellcolor{rowgray}49.04 & \cellcolor{rowgray}3.54 & \cellcolor{rowgray}15.12 \\ 
\multirow{2}{*}{never}
& \sf{FV} & 22.08 & 3.38 & 6.12 \\
& \cellcolor{rowgray}\sf{FRET} & \cellcolor{rowgray}51.33 & \cellcolor{rowgray}3.54 & \cellcolor{rowgray}15.12 \\ 
\multirow{2}{*}{within 3 seconds}
& \sf{FV} & 22.58 & 2.75 & 6.5 \\
& \cellcolor{rowgray}\sf{FRET} & \cellcolor{rowgray}79.88 & \cellcolor{rowgray}4.38 & \cellcolor{rowgray}23.25 \\ 
\multirow{2}{*}{for 3 seconds}
& \sf{FV} & 22.71 & 2.75 & 6.75 \\
& \cellcolor{rowgray}\sf{FRET} & \cellcolor{rowgray}76.71 & \cellcolor{rowgray}4.38 & \cellcolor{rowgray}23.0 \\ 
\multirow{2}{*}{after 3 seconds}
& \sf{FV} & 28.21 & 2.75 & 8.38 \\
& \cellcolor{rowgray}\sf{FRET} & \cellcolor{rowgray}121.88 & \cellcolor{rowgray}4.54 & \cellcolor{rowgray}35.17 \\ 
\multirow{2}{*}{until StopCondition}
& \sf{FV} & 26.83 & 3.75 & 8.12 \\
& \cellcolor{rowgray}\sf{FRET} & \cellcolor{rowgray}99.5 & \cellcolor{rowgray}4.67 & \cellcolor{rowgray}31.04 \\ 
\multirow{2}{*}{before StopCondition}
& \sf{FV} & 27.46 & 3.75 & 8.12 \\
& \cellcolor{rowgray}\sf{FRET} & \cellcolor{rowgray}111.58 & \cellcolor{rowgray}4.88 & \cellcolor{rowgray}33.33 \\ 
\bottomrule
\end{tabular}
\caption{Comparison of \Fret vs \textsc{Fv} (past semantics)}
\label{tab:FinSemDepthPast}
\end{table}
\end{minipage}
\end{table}

\subsection{Reflections on design choices}

Typically, each round of new formal verification or formalisation reveals new errors, inconsistencies, or choices that can be questioned. For example, the \NASA translation interprets the timed response \textit{eventually} in such a way that before the end of the trace the response MUST have taken place while a requirement using \textit{within n time units} is also true on a finite trace when there are less than $n$ time units before the end of the trace. We also found some other semantic choices that do make sense, but that can lead to counter-intuitive situations:
\begin{itemize}
    \item {\fret{}{}{TheParcel}{within 1 day}{satisfy BeDelivered}}$\not\Rightarrow${\fret{}{}{TheParcel}{eventually}{satisfy BeDelivered}}

    \item \fret{}{}{TheDriver}{after 3 hours of driving}{satisfy Rest} forces that the rest cannot occur before 3 hours of driving.
\end{itemize}

\section{Conclusions and future projects}\label{section:Conclusions}
We have presented the principle of Coherency consisting of (C1) and (C2) postulating that using various levels of formalisation is recommended and, moreover, that the different levels should exhibit similar logical structure. In light of this principle, we have given an alternative translation of \fretish to various fragments of MTL. The qualitative assessment is that, indeed, the \FV translation closely follows the description of the translation given by FRET. On the quantitative side, we see that on average, the new translation has certain benefits too.

However, since the new translation combines both past time and future time MTL operators, this currently prevents seamless integration with existing industrial verification workflows (see \cite{KatisEtAl:2025:StreamlinedReqBasedTesting, KatisEtAl:2022:RealizabilityCheckingInFRET, KooiMavridou:2019:RealizabilityCheckingInFRET, GiannakopoulouEtAl:2021:RealizabilityWithinFRET}), that currently do not support full MTL semantics. This highlights a trade-off between coherency and toolchain compatibility: while the new translation captures a more succinct and coherent representation of the original requirements, the current FRET translation benefits from tighter integration with mature verification back-ends, enabling end-to-end validation on large-scale models. In this sense, \cite{BogliEtAl:2025:NaturalFormalisations} would classify the \NASA translation as solution-oriented as opposed to problem-oriented where the latter would promote the problem to the highest level and from there, one has to choose the most suitable temporal logic and formalisation.

Both SNLs and CNLs are often met with enthusiasm: \cite{mueckstein:1985:CNL} speaks of the "the best of three worlds" (Natural language, Formal Language, and Dropdown Menu's) and the first line of the abstract of \cite{Giannakopoulou:2021:AutomatedFormalizationOfSNLReqs} reads: "The use of structured natural languages to capture requirements provides a reasonable trade-off between ambiguous natural language and unintuitive formal notations.". 
Indeed, the use of SNLs and CNLs are promising as taking the best of both worlds: the precision of formal languages on the one hand, combined with naturalness and perspicuity of natural languages on the other. Recently, more critical opinions have been expressed \cite{Idelberger:2023:SmartContractsAndCNLs} and our findings do indeed share those critical sentiments. However, we do believe that SNLs will play an important role in the advent of bridging LLMs to automated reasoning.

Indeed, it is a very natural idea to use CNLs as a way of reconciling the LLM paradigm with model checking, automated reasoning, and other logic-based techniques. Some preliminary promising results have been obtained \cite{HanhEtAl:2022:LLMsAndFormalLanguages, Liu:2022:LLM2LTL, MendozaEtAl:2024:NL2LTL, CoslerEtAl:2023:InteractiveNL2LTL} and in future research we wish to qualitatively and quantitatively investigate the effects of minor design changes in automated CNL translations on the performance of LLMs reasoning on the CNLs. 
Our hypothesis here is that Coherency will increase alignment between LLMs and automated reasoning tools.

\section*{Acknowledgements}
We thank Andreas Katis and Anastasia Mavridou for their patient explanations on the FRET tool and, we thank Guillermo Errezil for his encouraging support. Joosten has received financial support through the projects PID2023-151396OB-I00 and PID2023-149556NB-I00 of the Spanish Ministry of Science and Innovation, through 2021 SGR 00348 of the \textit{Generalitat de Catalunya} and, through the Severo Ochoa and María de Maeztu Program for Centers and Units of Excellence in R\&D (CEX2020-001084-M), the Spanish State Research Agency. We thank CERCA Programme/Generalitat de Catalunya for institutional support.


\end{document}